\documentclass{article}
\usepackage{spconf,amsmath,graphicx}
\usepackage{amssymb}

\usepackage{booktabs}
\usepackage{hyperref}
\usepackage{subcaption}
\usepackage{diagbox}

\usepackage{xcolor}

\title{Win the Lottery Ticket via Fourier Analysis: Frequencies Guided Network Pruning}

\name{Yuzhang Shang$^{1}$ \qquad Bin Duan$^{1}$ \qquad Ziliang Zong$^{2}$ \qquad Liqiang Nie$^{3}$ \qquad Yan Yan$^{1}$}

\address{$^{1}$Illinois Institute of Technology $^{2}$ Texas State University $^{3}$Shandong University\\
       }

\begin{document}
%
\maketitle
\begin{abstract}
With the remarkable success of deep learning recently, efficient network compression algorithms are urgently demanded for releasing the potential computational power of edge devices, such as smartphones or tablets. However, optimal network pruning is a non-trivial task which mathematically is an NP-hard problem. Previous researchers explain training a pruned network as buying a lottery ticket. In this paper, we investigate the Magnitude-Based Pruning (MBP) scheme and analyze it from a novel perspective through Fourier analysis on the deep learning model to guide model designation. Besides explaining the generalization ability of MBP using Fourier transform, we also propose a novel two-stage pruning approach, where one stage is to obtain the topological structure of the pruned network and the other stage is to retrain the pruned network to recover the capacity using knowledge distillation from lower to higher on the frequency domain. Extensive experiments on CIFAR-10 and CIFAR-100 demonstrate the superiority of our novel Fourier analysis based MBP compared to other traditional MBP algorithms.
\end{abstract}
\section{Introduction}
\label{sec:intro}
Over the past decade, utilizing deep learning to tackle intricate tasks, which are barely solved using shallow machine learning techniques solely, has achieved extremely outstanding performance in various domains of computer vision \cite{lecun2015deep, he2016deep,hu2021video}, information retrieval \cite{MMGCN,CLCRec} and multi-modal learning~\cite{zhu2021saying,zhu2020describing,hu2021coarse}. However, with the drastic increase of neural network parameters, computational resources are often short-handed. This hinders the broader deployment of deep models especially for edge devices. To solve this shortage problem, network pruning is proposed, which builds upon the prune-retrain-prune paradigm~\cite{lecun1990optimal} and is proved to be a practical approach to reduce the computation cost of overparameterized deep neural networks (DNNs).

Following the prune-retrain-prune paradigm, MBP is proposed in \cite{han2015deep, zhu2017prune} which show that pruning weights of small magnitudes and then retraining the model without contaminating the overall accuracy where simultaneously reach a pretty low compression rate on neural network models. Frankle et al.~\cite{frankle2018lottery} manifests that the MBP algorithm is to seek an efficient trainable subnetwork as a replacement for the original dense neural networks, where they empirically demonstrate that there exist sparse sub-networks that can be trained from scratch and perform competitively only if the initialization weights of this subnetwork are appropriate.
Moreover, MBP is studied further in \cite{Kusupati20, gale2019state} where the authors use a straight-through
an estimator in the backward pass and by which they achieve the state-of-the-art for unstructured pruning in DNNs. Yet despite all of MBP methods testify the effectiveness of MBP, no rational explanation has been endowed for why we can successfully retrain the pruned networks driven from MBP.


In this paper, we investigate the MBP scheme from a novel perspective, Fourier analysis where we regard DNN as a complex function extracting abstract representations for different datasets or tasks. Whilst pruning tries to remove less important representations, measuring the importance of those representations is non-trivial. Thanks to the power of Fourier analysis which can approximate any arbitrary functions by the sum of simpler eigenfunctions on the frequency domain, we intuitively want to measure the importance of the DNN's representations with Fourier analysis. It is proved that harmonic analysis, especially Fourier analysis, can efficiently obtain a spectral perspective of a perplexing system. However, due to the complexity of DNNs, conducting Fourier transformation on them remains a problem. Lately, \cite{xu2018understanding, yin2019fourier,raghu2017expressive} and \cite{rahaman2019spectral} study DNNs training by Fourier analysis and reveal some profound results which are: (i) DNNs tend to fit lower-frequency components first during training by gradient descent, and (ii) lower-frequency components are more robust to random perturbations of network parameters. We draw inspirations from~\cite{xu2018understanding, rahaman2019spectral} and design our network pruning scheme. Additionally, we endow an explanation using Fourier analysis for why the pruned network succeeds.

\begin{figure*}[ht]
    \centering
    \includegraphics[width=0.85\textwidth]{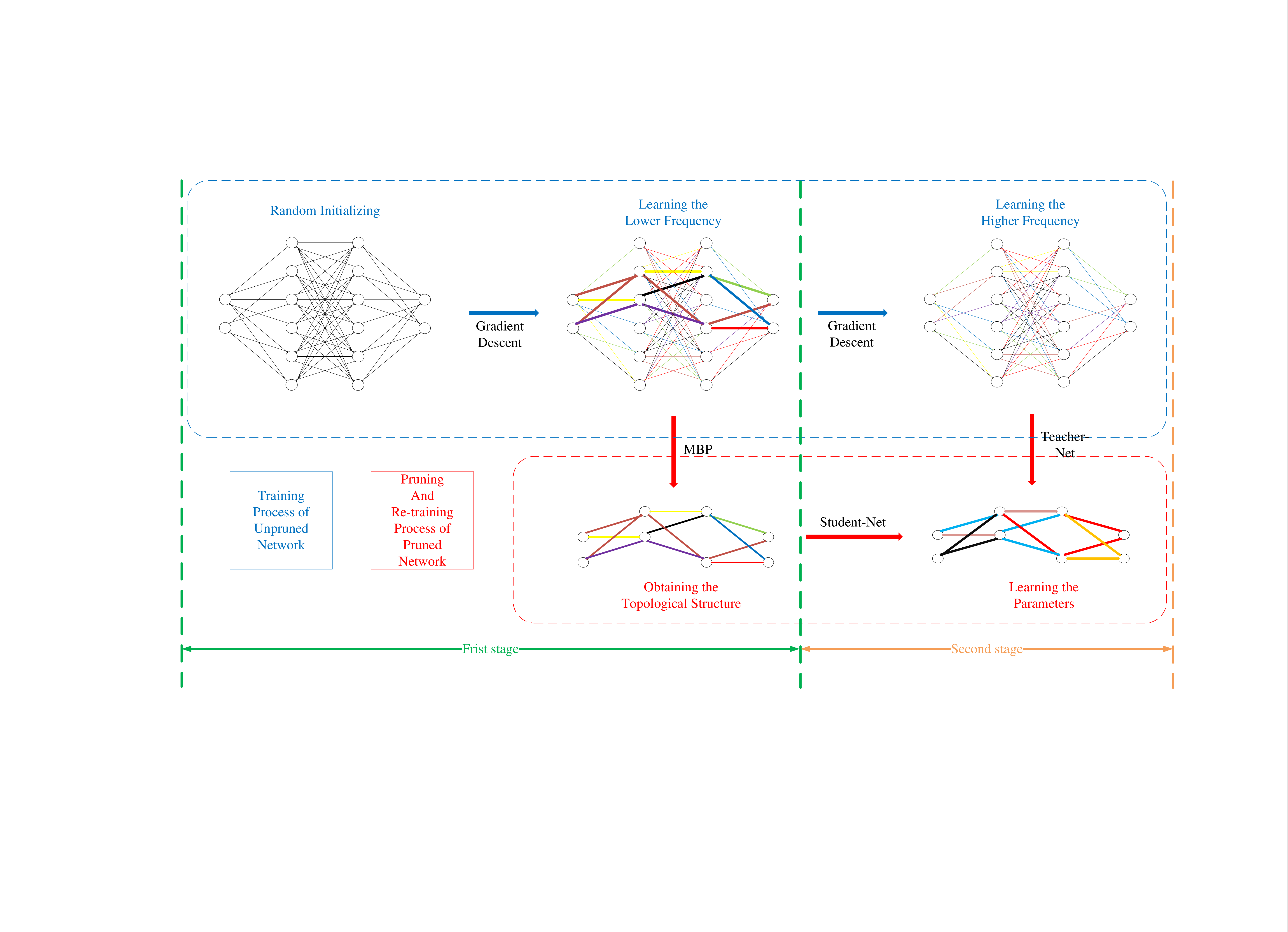}
    \captionsetup{font=footnotesize}
    \caption{The architecture of WILTON. The upper line demonstrates that the knowledge of data is divided into lower frequencies and higher frequencies in terms of the phase of training unpruned network by gradient descent. The bottom line shows that the two-stage pruning scheme firstly learns the topology structure from lower frequency and then use knowledge distillation to transfer the higher frequency to the pruned network.}
    \vspace{-0.2in}
    \label{fig:pipeline}
\end{figure*}

By embracing network pruning with Fourier analysis, the knowledge of data can be divided into lower frequencies and higher frequencies in the frequency domain in term of the phase of training unpruned network. Motivated by above assumptions, we propose a novel two-stage pruning method \textbf{WILTON} (\textbf{WI}n the \textbf{L}ottery \textbf{T}ickets through the F\textbf{O}urier A\textbf{N}alysis: Frequencies Guided Network Pruning, using the names “Lottery Tickets” adopted from “Lottery Ticket Hypothesis” \cite{frankle2018lottery}). We first acquire the topological structure and initialization weights by MBP scheme from the network representing lower-frequency, and then utilize knowledge distillation to recover the capacity where we learn the pruned network from the unpruned network which contains higher-frequency components. As shown in Fig.~\ref{fig:pipeline}, our architecture is designed in a parallel fashion to separately learn the structure and parameters of the pruned network.

In summary, our contributions can be highlighted as follows. (i) Fourier transform is introduced to the network pruning tasks to understand the generalization ability of MBP. (ii) A novel pruning method called WILTON is proposed to train the pruned neural network gradually from lower frequency to higher frequency. (iii) Our experimental results demonstrate that the proposed pruning scheme outperforms other recent MBPs on various network architectures, including ResNet \cite{he2016deep} and VGGNet \cite{simonyan2014very}.

\section{WILTON}
\label{sec:method}

\subsection{Fourier Analysis of Neural Networks}
For general purpose, we only take the networks with rectified linear unit (ReLU) activations as an example in this section.\par

We define the ReLU network with $L$ layers of widths $d_1,\cdots d_L (d = \sum_{k=1}^L d_k)$ and a single output neuron a scalar function$f : \mathbb{R}^d \longmapsto \mathbb{R}$ :
\begin{equation}
    f(\mathbf{x}) = (T^{(L)}\circ\sigma\circ T^{(L-1)}\circ \cdot\cdot\cdot \circ\sigma\circ T^{(1)})(\mathbf{x})
    \label{eq:1}
\end{equation}
where each $T^{(k)}: \mathbb{R}^{d_{k-1}} \longmapsto \mathbb{R}^{d_{k}}$ is an affine function and $\sigma(\mathbf{u}) = max(0, u_i)$ denotes the ReLU activation function acting elementwise on a vector $\mathbf{u} = (u_1,\cdots u_n)$. In the standard basis, $T^{(k)}(\mathbf{u}) = W^{k}\mathbf{u}+\mathbf{b}^{k}$, the $W^{k}$ and $\mathbf{b}^{k}$ stand for the weight matrix and bias vector of the network's $k$-th layer, respectively. For clarity and brevity, we discard the bias term of the network, then the network can be represented as: 
\begin{equation}
    f(W^1,\cdots,W^L;\mathbf{x})
    \label{eq:2}
\end{equation}
\par

The Fourier transform of the ReLU network is fairly an intricate mathematical problem. In \cite{diaz2016fourier} and \cite{rahaman2019spectral}, the authors develop an elegant procedure for evaluating it in arbitrary dimensions via a recursive application of Stokes theorem. We study the structure of ReLU networks in terms of their Fourier representation, $f(\mathbf{x}) := {(2\pi)^{d/2}\int\Tilde{f}(\mathbf{k})e^{i\mathbf{k}\cdot\mathbf{x}}\mathbf{dk}}$, where $\Tilde{f}(\mathbf{k}) := \int f(\mathbf{x})e^{i\mathbf{k}\cdot\mathbf{x}}\mathbf{dx}$ is the Fourier transform. We skip the formula derivation process and only present the main corollary here.\par

The Fourier components of the ReLU network $\widetilde{f}_\theta$ with parameters $\theta$ is given by the function:
\begin{equation}
    \widetilde{f}_\theta(\mathbf{k}) = \sum_{n=0}^{d}\frac{C_n(\theta,\mathbf{k})1_{H_n^{\theta}}(\mathbf{k})} {k^{(n+1)}}
    \label{eq:3}
\end{equation}
where $C_n(\theta,\cdot):\mathbb{R}^d \longmapsto \mathbb{C}$, $H_n^{\theta}$ is the union of n-dimensional subspaces and $1_{H_n^{\theta}}$ is the indicator over ${H_n^{\theta}}$. Note that Eq.\ref{eq:3} applies to general ReLU networks with arbitrary width and depth.\par

The numerator in Eq.\ref{eq:3} is bounded by $O(L_f)$, where $L_f$ is the Lipschitz constant of the network $f(\mathbf{x})$. Further the Lipschitz constant $L_f$ can be bounded by:
\begin{equation}
    L_f \leq \prod_{k=1}^{L}\Vert W^{k}\Vert \leq \Vert\theta\Vert_{\infty}^{L}\sqrt{d}\prod_{k=1}^Ld_k
    \label{eq:4}
\end{equation}
where $\Vert\cdot\Vert$ is the spectral norm of matrix and $\Vert\cdot\Vert_\infty$ is the max norm of vector, $d_k$ is the number of units in $k$-th layer, $d=\sum_{k=1}^Ld_k$ and $\theta$ is the concatenated vector of the network's whole parameters.

\subsection{WILTON Pruning Architecture}
For mathematical formulation, we use ${M}^k\in {\{0,1\}}^{d_{k-1}\times d_k}, (k = 1,\cdots L)$ to denote a mask on $W^k$ and define $W_p^k, (k = 1,\cdots L)$ as the parameters of pruned network. Note that the matrix ${M}^k, W^k$ and $W_p^k, (k = 1,\cdots L)$ have same shape. Then the network pruned network process can be defined by the matrix element-wise (Hadamard) product of mask $M^{k}$ and original weights $W^{k}$: $W_p^k = M^{k}\odot W^{k}, (k=1,\cdots L)$. The pruned network can be represented as:
\begin{equation}
    f(W_p^1,\cdots,W_p^L;\mathbf{x}) = f(M^{1}\odot W^{1},\cdots,M^{L}\odot W^{L};\mathbf{x}). 
    \label{eq:5}
\end{equation}
And then we can define the sparsity of pruned network $s = \frac{\sum_{k=1}^L\Vert M^k\Vert_0}{d}$.
A pruning strategy should yield a pruned network, which can maximize prediction accuracy given an overall parameter budget.

In the view of knowledge distillation, \cite{lee2018self,yim2017gift,shang2021lipschitz} demonstrate that the distilled knowledge can be considered as the flow between layers of network. As the neural network maps from the input space to the output space through many layers sequentially, by simplifying Eq.\ref{eq:2}, we assume the distilled knowledge to be transferred in terms of flow between layers, which is calculated by computing the Matrix-chain multiplication from successive layers' matrix:
\begin{equation}
    \mathbf{\bar{W}} = \prod_{k=1}^{L}W^{k}.
    \label{eq:6}
\end{equation}
Hereto, we define $\mathbf{\bar{W}}$ an important indicator to measure two networks' similarity. So, this indicator of the pruned and unpruned network should be close. In other word, after pruning the change form $\mathbf{\bar{W}}$ to $\mathbf{\bar{W}}_p = \prod_{k=1}^{L}W^{k}_p$ should be stable. And based on Eq.\ref{eq:4}, we have an approach to quantitatively analyze this term.


We obtain an upper bound of $\Vert\mathbf{\bar{W}}\Vert$:
\begin{equation}
    \Vert\mathbf{\bar{W}}\Vert = \Vert\prod_{k=1}^{L}W^{k}\Vert \leq \prod_{k=1}^{L}\Vert W^{k}\Vert.
    \label{eq:7}
\end{equation}

By combining Eq.\ref{eq:4} and Eq.\ref{eq:7}, we have:
\begin{equation}
    \Vert\mathbf{\bar{W}}\Vert \leq \prod_{k=1}^{L}\Vert W^{k}\Vert \leq \Vert\theta\Vert_{\infty}^{L}\sqrt{d}\prod_{k=1}^Ld_k.
    \label{eq:8}
\end{equation}

Therefore we only have to focus on the last term of Eq.\ref{eq:8}. After pruning the redundant weights from unpruned network, we have $\Vert\mathbf{\bar{W}}_p\Vert = \Vert\prod_{k=1}^{L}W^{k}_p\Vert$ of pruned neural network. The pruning scheme should have the ability to keep this term stable. Because pruning itself reduces the number of parameters of layers, which leads to the decrease of $d_k$ and $d$. Therefore, the pruning scheme must keep $\Vert\theta\Vert_{\infty}^{L}$ as big as possible. When pruning, we should prune those weights whose magnitudes are small to keep Eq.\ref{eq:8} stable, which is also an explanation of why magnitude-based pruning methods make sense from the perspective of Fourier analysis.

Besides the explanation, we also find more intriguing properties of DNN training through Fourier analysis. In \cite{xu2018understanding, rahaman2019spectral}, the authors both demonstrate that during the DNN training process, the lower frequency of data is learned first and more robust to parameters perturbation. Based on those properties, we manually unravel the lower and higher frequency knowledge of data on the frequency domain and bridge the ideology of knowledge distillation \cite{hinton2015distilling, ba2014deep}. We propose our two-stage pruning scheme, WILTON, where we first obtain a topological structure and initialization weights by MBP from a low-accuracy network and then distillate knowledge from the high-accuracy unpruned network into our pruned network. The architecture of proposed WILTON is shown in Fig.~\ref{fig:pipeline}. 


\section{experiment}
\label{sec:experiment}
\noindent\textbf{Datasets and Metrics.}~CIFAR-10 and CIFAR-100 \cite{krizhevsky2009learning} are widely used image classification datasets, which consists of 60k $32\times 32$ color images divided into 10 and 100 classes, respectively. We compare the top-1 accuracy on CIFAR datasets for our approach (WILTON), with vanilla MBP \cite{han2015deep}, LT \cite{frankle2018lottery} and STR \cite{Kusupati20}.

\noindent\textbf{Implementation details.}~We define the pruning ratio to be $r_p = s\cdot 100(\%)$, where $s$ is the sparsity of pruned network. For a given pruning ratio $r_p$, we train the unpruned network for 200 epochs and prune the unpruned network at epoch \#100. We discuss this hyperparameter in Section.~3.3. After pruning, we use the knowledge distillation to train the pruned network by the standard way. Specifically, we train the models using SGD with momentum of 0.9, batch size of 512, and the weight decay rate of 0.0005, unless state otherwise. The initial learning rate is set to 0.2 and decayed by 0.1 at every 20 or 40 epochs for CIFAR-10 and CIFAR-100, respectively. Additionally, we use the standard data augmentation (i.e., random horizontal flip and translation up to 4 pixels) for both the unpruned and pruned models. We keep 10\% of the training data as a validation set and use only 90\% for training.

\begin{table*}[!t]
\vspace{-0.4in}
\begin{minipage}{.49\textwidth}
    \centering
    \captionsetup{font=footnotesize}
    \caption{Test accuracy of pruned ResNet32 and VGG16 on CIFAR-10 dataset. The higher the better. 10.00$\pm$0.00 means the network cannot be trained and thus infer as random guessing.}
    \scalebox{0.77}
    {\begin{tabular}{l|cll}
\hline
Pruning ratio          & 90\%                                     & \multicolumn{1}{c}{95\%}                 & \multicolumn{1}{c}{98\%} \\ \hline
ResNet32(Full Network) & 92.30                                    & \multicolumn{1}{c}{-}                    & \multicolumn{1}{c}{-}    \\ \hline
random                 & \multicolumn{1}{l}{65.12$\pm$3.89}       & 18.87$\pm$2.24                           & 10.00$\pm$0.00            \\
vanilla-MBP\cite{han2015deep} & 91.17$\pm$0.06                           & 87.35$\pm$0.20                           & 87.64$\pm$0.28            \\
LT\cite{frankle2018lottery}   & 92.08$\pm$0.18                           & 92.78$\pm$0.04                           & 91.24$\pm$0.21            \\
STR\cite{Kusupati20}         & 93.22$\pm$0.11                           & \textbf{92.76$\pm$0.16}                  & 91.48$\pm$0.10            \\ \hline
WILTON(Ours)           & \textbf{93.24$\pm$0.13}                  & \textbf{92.79$\pm$0.11}                  & \textbf{91.64$\pm$0.09}            \\ \hline
VGG16(Full Network)    & 91.11                                    & \multicolumn{1}{c}{-}                    & \multicolumn{1}{c}{-}    \\ \hline
random                 & 43.26$\pm$0.34                           & 10.0$\pm$0.00                            & 10.00$\pm$0.00            \\
vanilla-MBP\cite{han2015deep}& \multicolumn{1}{l}{87.26$\pm$0.15}       & 41.17$\pm$0.33                           & 10.00$\pm$0.00            \\
LT\cite{frankle2018lottery}   & 92.50$\pm$0.31                           & \textbf{91.17$\pm$0.09}                           & 10.00$\pm$0.00            \\ \hline
WILTON(Ours)           & \textbf{92.88$\pm$0.03}                & \textbf{91.26$\pm$0.38}                & 10.00$\pm$0.00            \\ \hline
\end{tabular}}
\label{tabel:cifar10}
\centering
    \captionsetup{font=footnotesize}
    \caption{Test accuracy of pruned network from various training period of unpruned ResNet32 with different pruning ratios.}
    \scalebox{0.75}
    {\begin{tabular}{c|c|c|c|c|c}
    \toprule
    \diagbox{Ratio}{Epoch $\#$}  & 0          & 50         & 100         & 150        & 200        \\ \hline
    90\%         & 93.25 & 93.27 & \textbf{93.37} & 92.28 & 91.30 \\ 
    95\%         & 92.60 & \textbf{92.90} & 92.71 & 92.66 & 92.09 \\ 
    98\%         & 91.40 & \textbf{91.73} & 91.71 & 91.37 & 91.23 \\ \bottomrule
    \end{tabular}}
    \label{tabel:ablation}
\end{minipage}
\hfill
\begin{minipage}{.49\textwidth}
    \centering
    \captionsetup{font=footnotesize}
    \caption{Test accuracy of pruned ResNet32, ResNet56 and VGG16 on CIFAR-100 dataset. The higher the better. 1.00$\pm$0.00 means the network cannot be trained and thus infer like random guessing.}
    \scalebox{0.77}{
    \begin{tabular}{l|cll}
\hline
Pruning ratio          & 90\%                                     & \multicolumn{1}{c}{95\%}                 & \multicolumn{1}{c}{98\%} \\ \hline
ResNet32(Full Network) & 72.30                                    & \multicolumn{1}{c}{-}                    & \multicolumn{1}{c}{-}    \\ \hline
random                 & \multicolumn{1}{l}{6.30$\pm$0.21}       & 1.00$\pm$0.00                           & 1.00$\pm$0.00            \\
vanilla-MBP\cite{han2015deep} & 34.96$\pm$0.90                           & 27.49$\pm$0.21                           & 1.00$\pm$0.00            \\
LT\cite{frankle2018lottery}   & 70.75$\pm$0.18                           & 70.26$\pm$0.27                           & 65.52$\pm$0.15            \\
STR\cite{Kusupati20}         & \textbf{72.23$\pm$0.06}                           & \textbf{71.40$\pm$0.17}                  & 68.05$\pm$0.13            \\ \hline
WILTON(Ours)           & \textbf{72.31$\pm$0.26}                  & 70.39$\pm$0.02                  & \textbf{68.77$\pm$0.19}           \\ \hline
ResNet56(Full Network) & 72.32                                    & \multicolumn{1}{c}{-}                    & \multicolumn{1}{c}{-}    \\ \hline
random                 & \multicolumn{1}{l}{11.02$\pm$0.85}       & 4.59$\pm$0.23                           & 1.00$\pm$0.00            \\
vanilla-MBP\cite{han2015deep} & 49.71$\pm$0.28                           & 27.86$\pm$0.57                           & 1.00$\pm$0.00            \\
LT\cite{frankle2018lottery}   & 70.88$\pm$0.25                           & 70.17$\pm$0.48                           & 66.94$\pm$0.95            \\
STR\cite{Kusupati20}         & \textbf{72.40$\pm$0.74}                           & \textbf{71.83$\pm$0.55}                  & 68.58$\pm$0.22            \\ \hline
WILTON(Ours)           & \textbf{72.31$\pm$0.29}                  & \textbf{71.39$\pm$0.32}                  & \textbf{69.00$\pm$0.19}           \\ \hline
VGG16(Full Network)    & 71.09                                    & \multicolumn{1}{c}{-}                    & \multicolumn{1}{c}{-}    \\ \hline
random                 & 29.70$\pm$0.77                           & 1.00$\pm$0.00                            & 1.00$\pm$0.00            \\
vanilla-MBP\cite{han2015deep}& \multicolumn{1}{l}{65.38$\pm$0.36}       & 8.27$\pm$0.54                           & 1.00$\pm$0.00            \\
LT\cite{frankle2018lottery}   & \textbf{70.49$\pm$0.63}                           & 39.50$\pm$0.48                           & 1.00$\pm$0.00            \\ \hline
WILTON(Ours)           & \textbf{70.44$\pm$0.03}                & \textbf{45.29$\pm$0.60}                & 1.00$\pm$0.00            \\ \hline
\end{tabular}}
    \label{tabel:cifar100}
\end{minipage}
\vspace{-0.23in}
\end{table*}

\subsection{Results on CIFAR-10}
On CIFAR-10~\cite{krizhevsky2009learning}, we perform experiments on two standard network architectures -- ResNet32 and VGG16 where ResNet32 contains 0.47M parameters and VGG16 consists of 15.3M parameters. We compared our method with vanilla MBP \cite{han2015deep, zhu2017prune}, LT \cite{frankle2018lottery} and STR \cite{Kusupati20}. We also report the performance of a full network and a random pruning baseline. All the methods are compared extensively in the pruning ratio of 90\%, 95\%, 98\%.

Overall, Table.\ref{tabel:cifar10} shows that our method WILTON constantly outperform or evenly perform other pruning state-of-the-art methods in all pruning ratios with different network architectures. For further comparisons, we discuss the experimental results for ResNet32 and VGG16, respectively. For ResNet32, as pruning ratio increases, the performances of all methods drop in different levels. However, among all methods, WILTON tends to decay slowly which shows the robustness of our method. For VGG16, the result shows the same trend as of ResNet32 -- accuracy with WILTON deteriorates more slowly than other methods. 
\vspace{-0.15in}
\subsection{Results on CIFAR-100}
On CIFAR-100 \cite{krizhevsky2009learning}, we conduct experiment with three network architectures -- ResNet32, ResNet56 and VGG16. We compare our method with vanilla MBP \cite{han2015deep, zhu2017prune}, LT \cite{frankle2018lottery} and STR \cite{Kusupati20}. We also report the performance of a full network and a random pruning baseline. All the methods are compared in the pruning ratio of 90\%, 95\%, 98\%.

Our method achieves competitive and even better performance for all pruning ratios on different network architectures in Table.\ref{tabel:cifar100}. For ResNet32, as pruning ratio increases, the performances of all methods drop in different levels. However, among all methods, WILTON tends to decay slowly which shows the robustness of our method. For instance, the experiment about pruning ResNet32 shows that the model pruned by WILTON retain 69\% accuracy even with only 2\% remaining weights. In contrast, the model pruned with vanilla-MBP lose the ability of inference completely. It proves the observation that lower-frequency components are more robust to recover the capability of the network. For ResNet56, the pruned model with only 10\% parameters remaining can outperform its unpruned counterpart over 0.3\%, which indicates the overparameterization of ResNet. For VGG16, the result shows the same trend as of ResNet32 -- accuracy with WILTON deteriorates more slowly than other methods. While high-pruning-ratio pruning networks without shortcut structure tends to block the pathway for forward and backward propagation, the pruning of VGG16 (without shortcut) shows that the structure of VGG16 without 98\% parameters is destroyed so that all methods perform randomly guessing. An intuitive explanation is that the network approximating the lower-frequency components can be represented by a sparse network, so the pruned network driven from WILTON can keep effective to a large extent. 
\vspace{-0.2in}
\subsection{Pruning Phase Evaluation and Ablation Study}
WILTON is a two-stage pruning scheme, first obtaining a topological structure and initialization weights of the pruned network by MBP and then distilling knowledge from the high-accuracy unpruned network into the pruned network. On the unpruned ResNet32 training process on CIFAR-10, we prune 90\% weights of the unpruned network at different epoch (\#0, \#50, \#100, \#150 and \#200) and compare different pruned networks. The results are shown in Table.~\ref{tabel:ablation}. The performance for early pruning tends to be better than late pruning where certifies the robustness of lower-frequency components. However, extreme case like pruning in epoch \#0 degrades the performance since the network has not learn enough representations. That also gives us insights on how to choose the time to pruning our model such that we can recover the capacity of the unpruned model.
\section{conculsion}

\label{sec:conclusion}
In this paper, we propose a novel two-stage pruning method named WILTON which is an interpretable and versatile pruning approach based on Frequency domain decomposition. By separating knowledge from lower frequency to higher frequency on the frequency domain, WILTON  provides a novel pathway for exploring network pruning. Finally, our proposed WILTON serves as the stepstone to understand the optimization and initialization of deep neural networks.

\bibliographystyle{IEEEbib}
\bibliography{strings,refs}

\end{document}